\documentclass{IOS-Book-Article}

\usepackage[T1]{fontenc}
\usepackage{titlesec}
\titleformat{\paragraph}[runin]
{\bfseries}{\theparagraph}{1em}{}

\usepackage{mathptmx}
\usepackage{footmisc}
\usepackage{times}
\usepackage{float}
\usepackage{latexsym}
\usepackage{graphicx} 
\usepackage{booktabs} 
\usepackage{tcolorbox}
\usepackage{hyperref}
\usepackage{multirow} 
\usepackage{adjustbox}
\usepackage{amsmath}
\usepackage{enumitem}
\usepackage{subcaption}
\usepackage{caption}
\usepackage{soul}\setuldepth{article}

\def\hb{\hbox to 11.5 cm{}}

\begin{document}

\pagestyle{headings}
\def\thepage{}
\begin{frontmatter}              

\title{Embracing Diversity: A Multi-Perspective Approach with Soft Labels}

\author[A,C]{\fnms{Benedetta} \snm{Muscato}\textsuperscript{*}},  
\author[B,C]{\fnms{Praveen} \snm{Bushipaka}\textsuperscript{*}},  
\author[A]{\fnms{Gizem} \snm{Gezici}},  
\author[C]{\fnms{Lucia} \snm{Passaro}},  
\author[A]{\fnms{Fosca} \snm{Giannotti}}
and
\author[B]{\fnms{Tommaso} \snm{Cucinotta}}  


\address[A]{Scuola Normale Superiore}  
\address[B]{Scuola Superiore Sant'Anna}  
\address[C]{University of Pisa}


\begin{abstract}
Prior studies show that adopting the annotation diversity shaped by different backgrounds and life experiences and incorporating them into the model learning, i.e. multi-perspective approach, contribute to the development of more responsible models.
Thus, in this paper we propose a new framework for designing and further evaluating perspective-aware models on stance detection task,in which multiple annotators assign stances based on a controversial topic. We also share a new dataset established through obtaining both human and LLM annotations. 
Results show that the multi-perspective approach yields better classification performance (higher F1-scores), outperforming the traditional approaches that use a single ground-truth, while displaying lower model confidence scores, probably due to the high level of subjectivity of the stance detection task.
\end{abstract} 

\end{frontmatter}

\section{Introduction}
Recent advancements in the field of Natural Language Processing (NLP) have underscored the importance of annotator disagreement, redefining it as a meaningful source of information regarding the task, the data and the annotators \cite{sandri2023don}, rather than dismissing it as noise .
Research has shown that Large Language Models (LLMs) may exhibit biases that align with dominant Western perspectives \cite{davani2024d3code} exposing inequalities that could negatively impact underrepresented communities, whose voices are often drowned out by majority opinions ~\cite{santurkar2023whose}. As LLMs evolve alongside humans, aligning them with human preferences becomes a crucial aspect of their design process ~\cite{zhi2024beyond,muscato2024overview}. To address this challenge, ~\emph{Perspectivism}~\cite{basile2021toward}, a growing approach in NLP community, leverages disaggregated datasets, where all individual annotators label are included, to capture human disagreements, promoting inclusion of diverse viewpoints.
This new schema, rather than relying on aggregated labels- such as those obtained through majority voting- utilizes the diversity of human opinions, allowing models to learn from human disagreement ~\cite{uma2021learning, akhtar2021whose}, thereby avoiding the marginalization of minority voices. In line with the perspectivist paradigm, we propose a new framework to design and further evaluate the multi-perspective approach on stance detection, specifically about controversial and subjective topics. The main goal of this paper is to create perspective-aware by design models that incorporate human disagreement into the model learning phase in a more responsible way. 

\paragraph{Contribution}

Given the context, this study aims to explore the effectiveness of the multi-perspective approach. Specifically, we examine whether this methodology can enhance overall model performance and confidence.
Our contributions are as follows:

\begin{enumerate}

\item We introduce a new version of the stance detection dataset on controversial topics based on \cite{gezici2021evaluation}, augmented with document summaries and LLM annotations.
\item We employ two distinct methodologies: a baseline approach that utilizes aggregated labels and a multi-perspective approach designed to incorporate minority viewpoints by representing labels in a more nuanced manner i.e. soft labels.
\item We evaluate if the multi-perspective approach leads to improved model performance compared to relying solely on aggregated labels.

\end{enumerate}

\section{Related Work}
In this section, we set the foundation for our pipeline by combining insights on perspectivism, soft labels, and model uncertainty, while exploring how LLMs can act as annotators to capture diverse perspectives. 

\paragraph{Disaggregated datasets} 

In human-labeled datasets, annotations are typically gathered through crowdsourcing, where crowd workers on specific platforms, like MTurk \footnote{\url{https://www.mturk.com}} or Prolific \footnote{\url{https://www.prolific.com}}, are asked to provide their opinion on a given statement. In such contexts, especially when the task is subjective and no single ground truth may exist, crowd workers may disagree for various reasons, such as subjective bias or the ambiguity of the instance \cite{sandri2023don}. Consequently, recent studies have addressed this issue from a perspectivist standpoint, collecting each annotator's label to account for a range of diverse opinions, leading to the use of disaggregated datasets~\cite{basile2020s, leonardelli2023semeval}.
To promote this approach, the NLP community has recently released a list of publicly available perspectivist datasets\footnote{\url{https://pdai.info}}.

\paragraph{Learning from Soft Labels}

Soft labels provide a recent alternative to hard labels, i.e. aggregated labels, which are frequently criticized for oversimplifying complex data.
While one-hot encoding is used to assign a single, definitive value to each data point in hard labels, soft labels capture a range of possible values.
This renders the data more nuanced and better accounts for ambiguities and divergent viewpoints in annotations, reflecting the inherent uncertainty and variability in human judgment.
Previous studies modeled human diverse annotations
using soft labels \cite{peterson2019human, collins2022eliciting} achieving superior model performances improving also robustness and generalization \cite{pereyra2017regularizing}.


\paragraph{Model Uncertainty}

Different human judgments on subjective tasks introduce uncertainty into LLMs. The black-box nature of LLMs poses challenges in understanding how these models handle disagreements in text classification and generation~\cite{baan2023uncertainty}.~\cite{hu2023uncertainty} identify three sources of model uncertainty: the user input, the model architecture, and the final output. While accuracy is conventionally assessed with respect to the majority class, incorporating model uncertainty through representing multiple perspectives may be more beneficial.

\paragraph{Leveraging LLMs as annotators}
Lately, LLMs have demonstrated impressive capabilities in semantic understanding~\cite{chang2024survey, riccardi2023two}, as they easily interact with users, both in few-shot and zero-shot scenarios.
The current trend involves employing LLMs in various roles, such as acting as annotators to perform a wide range of tasks~\cite{gilardi2023chatgpt, mohta2023large}.
Although this procedure often requires substantial resources and domain expertise, cutting-edge LLMs like GPT-4~\cite{achiam2023gpt}, LLama-3~\cite{dubey2024llama} present viable substitutes, albeit with drawbacks of their own~\cite{pavlovic2024effectiveness}. Several works have explored the labeling capabilities of LLMs for subjective tasks, including stance detection, hate speech detection, and narrative analysis~\cite{zhu2023can, ziems2024can}.
While these models are frequently fine-tuned to align with human preferences, current research have investigated whether these models accurately represent human disagreements~\cite{lee2023can, pavlovic2024effectiveness}. 
The cost of using LLMs as annotators is a significant advantage over employing humans, but it is important to recognize that they may introduce biases~\cite{wang2024human}.

\vspace{1.5em}
\begin{figure*}[!ht]
    \centering
    \vspace{-1em}  
    \includegraphics[width=\linewidth]{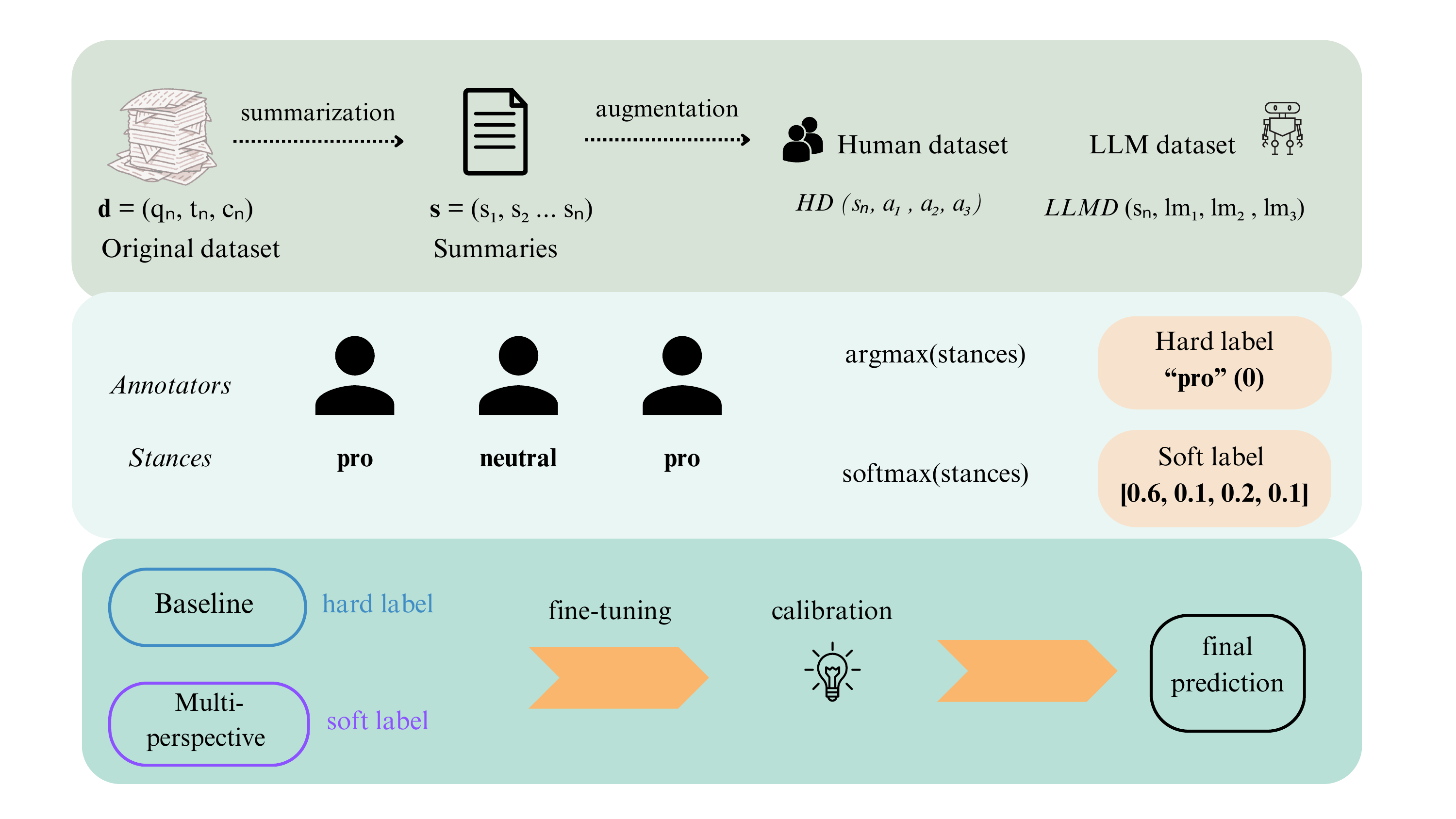} 

        \vspace{-0.5em}  
        \caption{The multi-perspective framework for the stance detection task includes the dataset preparation phase with the summarization and further augmentation steps via obtaining LLM annotations. Then, annotations are transformed into hard and soft labels, model fine-tuning is fulfilled and classifier's final prediction scores are calibrated.}
        
      \label{fig:fig1}
    \vspace{1.5em}  
\end{figure*} 

\vspace{1em}
\section{Methodology}
\label{sec:metodology}
We propose a multi-stage framework specifically for the stance detection task, as illustrated in Figure~\ref{fig:fig1}. The stance detection task which aims to automatically determine the writers’ stance (perspective, or viewpoint) towards a target in a given textual content. In the scope of this study, the target is a claim about the corresponding controversial topic and the author may support the claim with a stance of~\emph{pro}, disagree with it by taking the stance of~\emph{against}, or choose not to have a clear stance which is~\emph{neutral}, i.e. neither agree, nor disagree with it. Thus, the stance can take three different values as~\emph{pro, against, or neutral} towards a given claim, and additionally~\emph{not-about} which means that the author's statement is not relevant with the given claim (target). 
To investigate the effect of the perspectivist approach on stance detection task, we use two different methodologies:~\emph{Baseline} model with hard labels and~\emph{Multi-Perspective} model with soft labels (the overview of these methodologies can be found in Figure \ref{fig:fig1}).

The proposed pipeline\footnote{This framework is adaptable to different applications, for example summarization can be replaced with paraphrasing if necessary.} consists of the following steps:
\textit{Step 1} is the summarization of the documents in the original dataset, \textit{Step 2} is the data augmentation via obtaining LLM annotations to create two distinct datasets as human (HD) and LLM dataset (LLMD) and \textit{Step 3} is the model fine-tuning, along with the calibration (Section~\ref{sec:experiments} for more details).

\subsection{Baseline Model}
\label{sec:baseline}
In traditional machine learning settings, label aggregation techniques such as~\emph{majority voting} are typically applied to create a single label for each data instance. In the baseline model, we follow the traditional paradigm in which the majority label that is the most frequent label among the multiple annotations provided by the annotators is created and used for each data instance. Majority labels are aggregated using~\emph{hard labels} that are in traditional binary classification settings encoded as $0$ or $1$. In our multi-class scenario, we refer to the majority label as the index of the most common option, represented as the hard label. Each index corresponds to a specific stance label in the following order:~\emph{pro} (0),~\emph{against} (1),~\emph{neutral} (2), and~\emph{not-about} (3). An example about data transformation is illustrated in Figure~\ref{fig:fig1}.

\subsection{Multi-Perspective Model}
\label{sec:multip}
In the multi-perspective approach, unlike the baseline, a majority label is not generated, instead the multi-perspective model uses disaggregated labels. These disaggregated labels initially represented as discrete values, are converted into continuous values through a softmax function namely \emph{soft labels} ~\cite{uma2020case}. 
The advantage of using soft labels is that they represent a probability distribution over the possible classes, which can enhance the model performance, particularly in subjective tasks where annotator choices may differ significantly~\cite{plank2022problem}.
Since the baseline and multi-perspective approaches handle the dataset design differently, the multi-perspective approach applies the soft loss~\cite{uma2020case} instead of the standard cross entropy loss. This choice stems from the need to represent the distribution of human labels in a more nuanced way. The soft loss is defined as:
\vspace{-1.5em}
\begin{equation*}
-\sum_{i=1}^n \sum_{c} p_{\text{hum}}(y_i = c \mid x_i) \log p_\theta(y_i = c \mid x_i)
\end{equation*}

where $p_{\text{hum}}(y \mid x)$ represents the human label distribution (i.e. soft labels) which is obtained by applying the softmax function to the logits produced by the classifier.

\section{Experimental Setup}
\label{sec:experiments}
This section outlines the technical details of the conducted experiments. Our code and results are publicly available at \url{https://anonymous.4open.science/r/perspectivism-0473}. We first describe the overall pipeline as displayed in Figure~\ref{fig:fig1}.
The original dataset texts (already provided with human hard labels) were first summarized and then further augmented with additional annotations from LLMs. This process resulted in the creation of two distinct datasets: HD, containing the original human-derived annotations, and LLMD, incorporating annotations generated by LLMs. Then, these annotations were converted into hard labels, for the Baseline, and soft labels for the Multi-Perspective model, i.e. with the aim of representing the diverse perspectives in a more fine-grained manner. Subsequently, model fine-tuning was performed, and classifier predictions were calibrated.

\subsection{Original Dataset}
\label{subsec:originaldataset}
 
For this study, we built upon the work of ~\cite{gezici2021evaluation}. The dataset comprises the top 10 news search results retrieved from Google and Bing in response to 57 queries on controversial topics. These topics range from education, health, and entertainment to religion and politics, all of which are known to evoke diverse perspectives and opinions. Controversial issues in these domains often touch on deeply held values, beliefs, and societal debates, making the dataset particularly suitable for a study rooted in perspectivism, where differing viewpoints are central to the analysis.
Each dataset instance is composed of a query (about a controversial topic), document title and the textual content with respect to a given query of the varying lengths, i.e. $1.200$ tokens on average and extend up to $7.000$ tokens. 
Each document (title and content) with respect to the given query has been annotated by three annotators on MTurk (Figure~\ref{fig:fig1}). We use this stance detection dataset after the data augmentation steps below to apply our proposed multi-perspectivist approach in the scope of subjective tasks. The reported Fleiss-Kappa\footnote{Fleiss Kappa is a statistical measure of agreement which is an extended version of Cohen's Kappa (only for two raters) that takes into account the agreement due to chance as well.} score of $0.35$ and inter-rater agreement score of $0.49$ on the original dataset highlight the subjectivity and ambiguous nature of the stance detection task.
Note that each document has not been annotated by the same three annotators due to the design choices which leads to a more enriched dataset with diverse opinions (annotations).

\newtcolorbox[auto counter, number within=section]{mybox}[2][]{%
    colback=blue!5!white,
    colframe=black!75!black,
    fonttitle=\bfseries,
    title=#2,
    before=\vspace{-5pt}, 
    after=\vspace{-5pt},
    #1
}

\begin{figure}[!t]
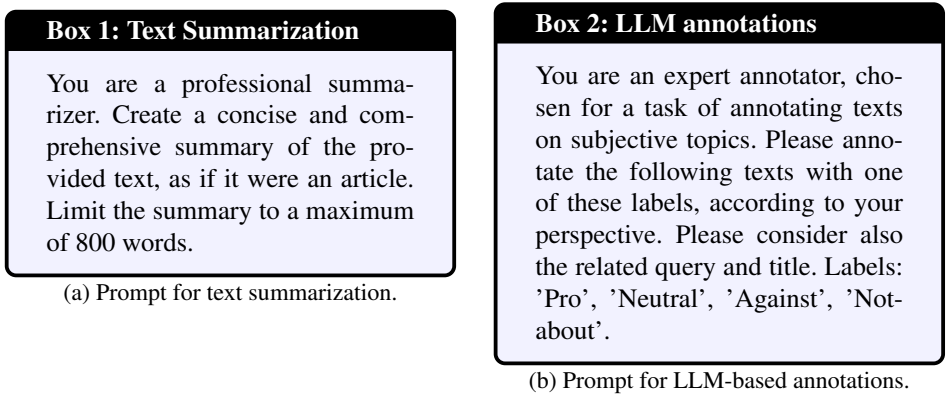

\centering
\begin{subfigure}[b]{0.48\textwidth}
\begingroup
\sbox0{\begin{minipage}{\linewidth}
\begin{mybox}{Box 1: Text Summarization}
You are a professional summarizer. Create a concise and comprehensive summary of the provided text, as if it were an article. Limit the summary to a maximum of 800 words.  
\label{fig:box_summarization} 
\end{mybox}
\caption{Prompt for text summarization.}
\end{minipage}}
\raisebox{\dimexpr.5\ht0-.5\height}{\usebox0}
\endgroup
\end{subfigure}%
\hfill
\begin{subfigure}[b]{0.48\textwidth}
\begingroup
\sbox0{\begin{minipage}{\linewidth}
\stepcounter{tcb@cnt@mybox} 
\begin{mybox}{Box 2: LLM annotations}
You are an expert annotator, chosen for a task of annotating texts on subjective topics. Please annotate the following texts with one of these labels, according to your perspective. Please consider also the related query and title. Labels: 'Pro', 'Neutral', 'Against', 'Not-about'.
\label{fig:box2_annotations} 
\end{mybox}
\caption{Prompt for LLM-based annotations.}
\end{minipage}}
\raisebox{\dimexpr.5\ht0-.7\height}{\usebox0}
\endgroup
\end{subfigure}
\caption{GPT4-Turbo prompts for text summarization (a) and for LLM annotations (b).}
\label{fig:prompt_comparison}
\end{figure}

\subsection{Dataset Augmentation}
\label{sec:data_augment}

\paragraph{Summarization}
\label{summarization}
Since the original dataset contains long documents with varying lengths, we decided to first apply summarization due to the maximum input length of transformer-based models, namely BERT~\cite{devlin2018bert} and RoBERTa~\cite{liu2019roberta}. Both of these models have the maximum input length of $512$ and summarization (instead of truncation) can provide more enriched contents especially for the long documents.
We applied summarization only on those documents (title and content) longer than $800$ tokens (empirically determined) since summarization might lead to information loss on shorter documents. 

In our initial summarization experiments, we employed various models, including Pegasus-CNN-DailyMail, BART-large-CNN, and Falcon-7b-Instruct. 
To assess the quality of these summaries, we compared them using the ROUGE score, as shown in Table~\ref{tab:rouge_scores1}, Table~\ref{tab:rouge_scores2} and Table~\ref{tab:rouge_scores3}. 

\paragraph{Pegasus-CNN-DailyMail}
Table \ref{tab:rouge_scores1} shows the summarization performance of Pegasus-CNN-DailyMail. The model exhibits a high precision but low recall and F1 scores across ROUGE metrics, indicating limited content coverage.

\paragraph{BART-large-CNN}

BART-large-CNN outperforms Pegasus-CNN-DailyMail (Table \ref{tab:rouge_scores2}), with higher recall and slightly lower precision scores. Better recall scores indicate that BART-large-CNN retrieves the content more effectively, while slightly lower precision scores suggests a trade-off, as the summaries may include less relevant details.

\paragraph{Falcon-7B-Instruct}

Falcon-7B-Instruct is aligned with the precision and recall results of BART-large-CNN, indicating a comparable performance in summarization. These similar results suggest that Falcon-7B-Instruct also effectively balances content coverage and relevance. 
\paragraph{GPT-4-turbo}
\label{par:sumgpt}
Based on the automated evaluation and human evaluation of the generated summaries of the aforementioned models, we decided to use GPT-4-Turbo via OpenAI batch API \footnote{Note that due to the maximum input length constraints of these models, we chunked the input text to be fed into these models except GPT-4 turbo\footnote{\url{https://openai.com/index/gpt-4/}} owing to its larger maximum input length.}. 
For the detailed comparative evaluation results we leveraged different metrics, namely ROUGE, i.e. measuring the overlap of n-grams between the generated summary and the reference, using an automated package~\cite{lin-2004-rouge}, BERTScore~\cite{zhang2019bertscore}, i.e. semantic similarity, and BLEU score~\cite{papineni2002bleu}, i.e. n-gram similarity.
Overall model evaluation results display that the GPT-4 Turbo model has a moderate performance on the ROUGE score, high performance on the BERTScore, whereas it shows a low performance on the BLEU score. Based on these results, we decided to use GPT-4 Turbo for the summarization phase, since BLEU score is typically used for machine translation and not suitable for summarization tasks.
The evaluation of the GPT-4 summaries is presented in Table~\ref{tab:tabgpt}.

%
Since prompting is known to highly affect the model performance on a wide range of tasks\footnote{\url{https://platform.openai.com/docs/guides/prompt-engineering/strategy-write-clear-instructions}}, we conducted various experiments with prompt engineering for two main purposes: first to improve summarization using the GPT-4-turbo model, and second, to obtain LLM-based annotations. Two selected prompts after experimenting with different prompts \footnote{For LLM annotations collection we designed a prompt in zero-shot settings to minimize any potential bias} are illustrated, respectively, in Figure \ref{fig:prompt_comparison} . 
Considering the overall performances of the aforementioned models followed by a human evaluation (due to the repetition of information in previous summaries), we decided to switch to GPT-4. GPT-4-turbo exhibits a modest precision score, yet low recall and F1 scores. However, high BERT-score suggests that the generated summaries are semantically relevant. As expected, the BLUE score is low due to matching n-grams, where word-to-word matching is not typically required.

\begin{table}[h!]
\centering
\scalebox{1.0}{ 
\begin{tabular}{lccc}
\toprule
\textbf{Metric} & \textbf{Precision} & \textbf{Recall} & \textbf{F1 Score} \\
\midrule
\textbf{ROUGE-1} & 0.4998 & 0.1774 & 0.2549 \\
\midrule
\textbf{ROUGE-2} & 0.1770 & 0.0520 & 0.0779 \\
\midrule
\textbf{ROUGE-L} & 0.4571 & 0.1620 & 0.2329 \\
\midrule
\textbf{BERT-Score} & \textbf{0.8503} & 0.8357 & 0.8429 \\
\midrule
\textbf{BLEU Score} & 0.0171 & - & - \\
\bottomrule
\end{tabular}
}
\vspace{0.9em}
\caption{ROUGE score on GPT-4-turbo's summaries}
\label{tab:tabgpt}
\end{table}

\paragraph{Augmentation via LLM annotations}
\label{augment}

To establish models that are more responsible and with a higher capability of representing diverse perspectives, we experimented with three different LLMs.
We opted for open-source SOTA models, in particular LLama-3-8b~\cite{dubey2024llama}, Mistral-7b~\cite{jiang2023mistral} and Olmo-7b~\cite{groeneveld2024olmo}\footnote{We selected the listed LLMs based on their availability and GPU constraints. The models were loaded onto the GPU using half-precision (float16).}.

Based on the label distribution obtained by the LLMs on the training, validation, and test sets, the~\emph{pro} is the most frequently assigned stance label by all three LLMs (37\%, 50\%, and 69\% for Olmo, LLama-3, and Mistral respectively), while Olmo exhibits a significantly higher percentage for the~\emph{against} (37\% vs. 22\% and 15\%). The four label distribution charts including the LLM with majority vote can be found in Figure~\ref{fig:fig2} and Figure~\ref{fig:fig3} in Appendix~\ref{sec:applabels}.

Furthermore, the percentage of full percentage of agreement, i.e. defined as the proportion of the cases where all annotators concur exactly on the same label, is quite low with 11\%. The low agreement score suggests a high level of disagreement among the LLMs, probably due to the difficulty of the annotation task and subjective nature of the dataset, i.e. which discusses debatable controversial topics. Similarly, Cohen's Kappa scores that were calculated for all LLM annotations in a pairwise manner also confirm low level annotator agreement.

Before summarization, each instance $d_i$ consists of $\{q_i, t_i, c_i\}$. From this, we first concatenated the document title $t_i$ and content $c_i$ and summarized it into $s_i$ (Step 1, as detailed in Section \ref{summarization}). After obtaining the summary $s_i$, we designed two distinct datasets: (i) the Human Dataset (HD), which consists of $HD = \{s_i, a_1, a_2, a_3, maj\}$, where $a$ represents crowd-annotated labels (obtained via crowdsourcing) and $maj$ indicates the majority labels, as described in Section~\ref{sec:baseline}; and (ii) the LLM-Annotated Dataset (LLMD), defined as $LLMD = \{s_i, lm_1, lm_2, lm_3, maj_{lm}\}$. The structure of LLMD is identical to HD, with the difference being that the annotations $lm$ are obtained from LLMs rather than human annotators, with the majority label $maj_{lm}$ similarly derived from LLM annotations.

\begin{figure*}[!h] 
    \centering
    \includegraphics[width=1.0\textwidth]{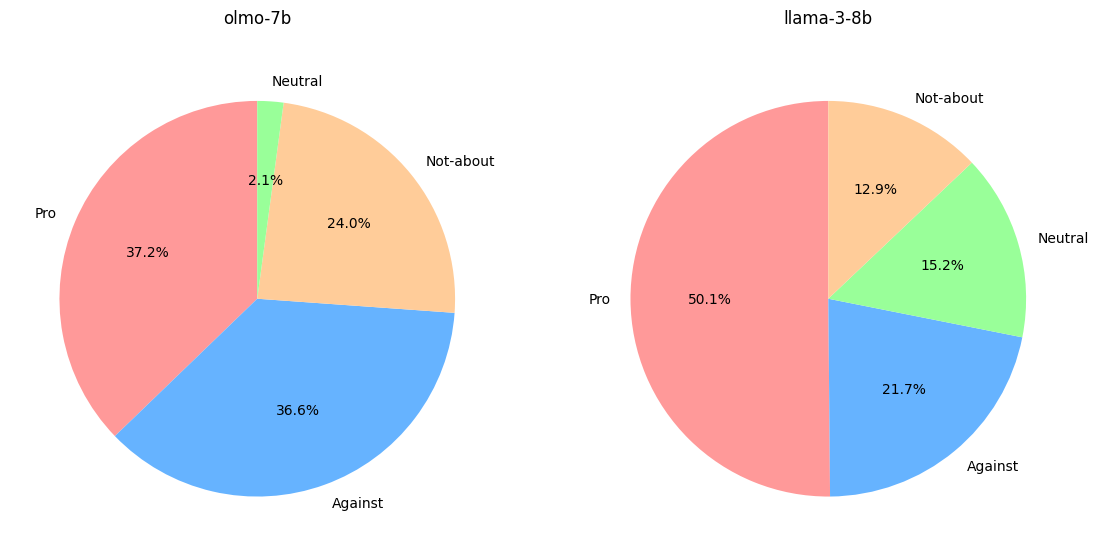}
    \caption{Olmo-7b and Llama 3-8b label distribution (train, val, test)}
    \label{fig:fig2}
\end{figure*}

\begin{figure*}[!h] 
    \centering
    \includegraphics[width=1.0\textwidth]{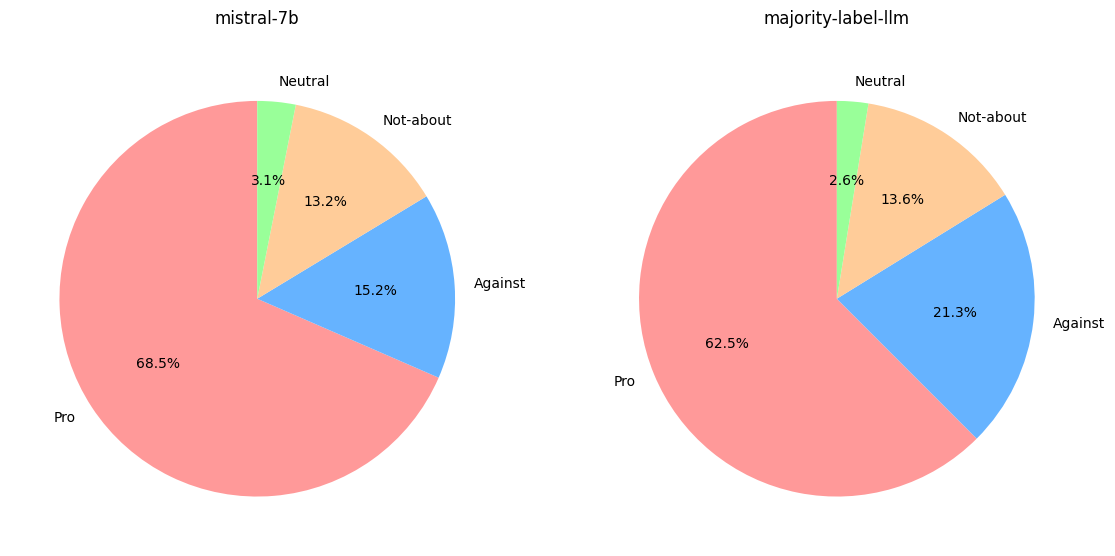}
    \caption{Mistral-7b and Majority label LLM distribution (train, val, test)}
    \label{fig:fig3} 
\end{figure*}

\newpage
\begin{table*}[!t]
\vspace{-1.5em}
\centering
\small 
\renewcommand{\arraystretch}{0.5} 
\adjustbox{max width=\textwidth}{ 
\begin{tabular}{clcccccc}
\toprule
\textbf{Approach} & \textbf{Dataset} & \textbf{Model} & \textbf{Acc.} & \textbf{Prec.} & \textbf{Rec.} & \textbf{F1} & \textbf{Avg. Conf.}\\ 
\midrule

\multirow{4}{*}{Baseline} & \multirow{2}{*}{HD} & BERT-large & 36.69 & 39.03 & 35.93 & 33.80 & 40.20 \\
& & & & & & & \\
& \multirow{2}{*}{HD} & RoBERTa-large & 56.11 & 61.11 & 58.04 & \textbf{57.22} & 57.25 \\
& & & & & & & \\
& \multirow{2}{*}{LLMD} & BERT-large & 60.78 & 15.50 & 24.60 & 19.01 & \underline{60.59} \\
& & & & & & & \\
& \multirow{2}{*}{LLMD} & RoBERTa-large & 61.76 & 15.44 & 25.0 & 19.09 & 60.44 \\
& & & & & & & \\
\midrule
\multirow{4}{*}{Multi-Perspective} & \multirow{2}{*}{HD} & BERT-large  & 46.76 & 46.88  & 47.16  & 46.75  & 45.82 \\ 
& & &  &  &  &  &  \\ & \multirow{2}{*}{HD} & RoBERTa-large & 60.43 & 63.55  & 62.83  & \textbf{61.90}  & \underline{48.76} \\
& & &  &  &  &  &  \\ 
& \multirow{2}{*}{LLMD} & BERT-large & 61.76  & 15.44 & 25.0  & 19.09  &  30.42 \\ 
& &  &  &  &  &  &  \\& \multirow{2}{*}{LLMD} & RoBERTa-large & 61.76  & 15.44 & 25.0  & 19.09  & 30.13  \\ & &  &  &  &  &  & \\
\bottomrule
\end{tabular}
}
\caption{Comparative evaluation results of different approaches and models.}
\label{tab:results}
\vspace{-1em}
\end{table*}

\paragraph{Dataset Preprocessing}
\label{sec:appdataset}
The dataset contains 1026 instances in total. Then, we applied the preprocessing steps to clean the dataset. We removed the null documents, then the ones with link-broken label, i.e. that are not accessible, and finally the documents without any majority label, i.e. all annotator disagree on the ground truth.
After removing the instances with no majority label, 897 instances left and the dataset was then splitted as 619 for training, 139 instances for validation and 139 instances for testing. This the actual dataset size of HD. Instead, LLMD contains 704 instances in total as 505 instances for training, 97 instances for validation, and 102 for testing. The reason of the dataset size differences of HD and LLMD results from the fact that there were more instances with no majority label based on the LLM annotations. Thus, to make a comparative evaluation of the baseline and multi-perspective approaches, we had to discard more instances for the LLMD.

\subsection{Model Learning}
\label{sec:model_learning}
We prepared the dataset for fine-tuning by combining the $q_i$ and $s_i$ of each dataset instance as input for both the baseline model (Section~\ref{sec:baseline}) and the multi-perspective model (Section~\ref{sec:multip}).
Notably, while the baseline model uses the majority label as the ground truth, the multi-perspective model leverages label probabilities. 

\subsubsection{Fine-tuning}
\label{finetuning} 
We fine-tuned the aforementioned LLMs using two different approaches as the baseline, relying on the conventional use of aggregated labels with majority voting, and the multi-perspective model, incorporating diverse perspectives, i.e. without transforming them into a single label, via soft labels to refine the model's learning process. 
For both of these approaches, we fulfilled model fine-tuning on 2 x 32GB Tesla V100s.

%

To fine-tune BERT-large and RoBERTa-large with baseline and multi-perspective approaches, we used default hyperparameters. We fine-tuned the both models for 6 epochs, with a learning rate of $1 \times 10^{-15}$, $weight$ $decay$ of $0.01$ and $500$ $warmup$ $steps$ and a batch size of 8.

\paragraph{Baseline Loss}

The loss function used for the baseline model is multi-class cross-entropy defined as:
\[
\text{Loss} = - \sum_{i=1}^C y_i \log(p_i)
\]
where $C$ is the number of classes, $y$ is a one-hot encoded vector representing the true class and $p$ is a vector of predicted probabilities. The goal is to minimize the cross-entropy loss during the training process which penalizes the model more when the predicted probabilities deviate from the true distribution of classes.

\vspace{1.0em}
\paragraph{Baseline}
The baseline approach shows a lower ECE of $0.03$ and $0.05$ with HD for the BERT and RoBERTa-large, meaning that the model is well-calibrated, i.e. small deviation from the perfect calibration.
On LLMD, the ECE is much higher as $0.35$ and $0.37$, suggesting that calibration made the uncalibrated models deviate more from the perfect calibration, i.e. model became over- or under-confident on its dataset predictions which signals a ill-calibrated model.

\paragraph{Multi-Perspective}

BERT-large performed with an ECE of $0.17$ on HD, which is significantly higher than the baseline value of $0.03$. This suggests that when BERT-large uses the multi-perspective approach, its calibration worsens on HD, as expected according to the findings in section \ref{calibration}.
However, with LLMD, the ECE drops dramatically to $0.05$, showing improvement in calibration. In this case, BERT-large seems to make much more reliable predictions according to calibration parameters.
Regarding RoBERTa-large: on HD, the ECE increases to $0.30$, which is higher than the baseline. This indicates that, like BERT-large, RoBERTa-large performs worse on HD using the multi-perspective approach, becoming less calibrated and thus less reliable according to traditional standards.
On LLMD, however, the scores improve to $0.05$, showing good calibration for RoBERTa-large as well.
In summary, the results indicate a trade-off when applying the multi-perspective approach: it improves model performance on LLMD (with lower ECE values) but leads to worse performance on HD (with higher ECE values). This pattern is consistent across both BERT-large and RoBERTa-large models.

\subsubsection{Calibration} 
\label{calibration}
After the fine-tuning step, we further applied calibration
to adjust the predicted scores from a classifier to better align with the true probabilities which can lead to a fairer comparative evaluation, particularly for the model confidence scores. As a calibration method, we employed temperature scaling~\cite{guo2017calibration}.

At its core, temperature scaling involves dividing the logits by a small value $T$ and then applying the softmax function to convert the logits ($z$) into a probability distribution over the possible outputs. The value $T$ is a hyperparameter often tuned on the validation set to minimize specific metrics e.g. negative log-likelihood, and we tuned the $T$ on our validation set for 6 epochs.

\section{Results}

The model evaluation results are reported using various metrics of accuracy, precision, recall, and F1 score alongside average model confidence scores on the test set in Table~\ref{tab:results}. Based on the results, multi-perspective models outperform the baseline models in most cases, except the baseline RoBERTa-large model on LLMD, which showed a similar performance with the Multi-Perspective BERT-large and RoBERTa-large models on LLMD. The results confirm that using soft labels improve the model performance.

The best-performing baseline model is RoBERTa-large fine-tuned on HD with the F1-score of $57.22$, while the best multi-perspective model is RoBERTa-large fine-tuned on HD with $61.90$. Nonetheless, both for the baseline and multi-perspective, HD models show superior performance in comparison to the LLMD models which reflects that humans provide annotations with higher quality in comparison to LLMs.
Apart from these, we can observe that baseline models generally exhibit higher model confidence scores (except the BERT-large model on HD) irrespective of the fine-tuning dataset (HD or LLMD). 
This is probably because the multi-perspective approach introduces higher level of model uncertainty through representing different viewpoints with equal weights. As a result, we argue that confidence score alone may not be the best criterion for evaluating multi-perspective models.

\vspace{0.5em}
\begin{table}[h!]
\centering
\vspace{-0.5em}
\begin{adjustbox}{width=0.7\textwidth} 
\begin{tabular}{lcc}
\toprule
\textbf{Approach} & \textbf{HD}& \textbf{LLMD} \\
\midrule
\textbf{Baseline} & & \\
\midrule
BERT-large & 0.04 (same) & 0.35 (same) \\
RoBERTa-large & \underline{0.04} ($U$), 0.06 ($C$) & 0.37 (same) \\
\midrule
\textbf{Multi-perspective} & & \\
\midrule
BERT-large & \textbf{\underline{0.05}} ($U$), 0.18 ($C$) & 0.20 ($U$) \underline{0.05} ($C$) \\
RoBERTa-large & \textbf{\underline{0.12}} ($U$), 0.30 ($C$)  & 0.18 ($U$), \underline{0.05} ($C$) \\
\bottomrule
\end{tabular}
\end{adjustbox}
\vspace{0.7em}
\caption{ECE (Expected Calibration Error) values
with \& without calibration denoted as $U$ and $C$ respectively, if the values are different.}
\label{table:ece_grouped}
\end{table}

The secondary focus of this paper is verifying whether model calibration ensures that the estimated class probabilities align closely with the actual outcomes.
After applying calibration, we aimed to measure 
if a given model is well-calibrated, through calculating Expected Calibration Error (ECE)~\cite{naeini2015obtaining} which can be used to quantify how well the predicted output probabilities of the model matches the actual probabilities of the ground truth distribution. Table~\ref{table:ece_grouped} shows that the uncalibrated baseline models are already well aligned with the perfectly calibrated model (ideal case with ECE of 0 since the lower ECE is, the better), thus calibration did not create a significant effect, while the situation is different for the multi-perspective approach. Calibration on the multi-perspective HD models lead to ill-calibrated models (higher ECE with calibration), while for the LLMD models, calibration helped to
decrease the deviation from the perfect calibration (lower ECE with calibration). Based on these results, we applied calibration on all the models except the models with the multi-perspective approach on HD.
While the calibration did not affect the overall model performance based on the evaluation metrics (accuracy, precision, recall, and F1 score), it significantly affected the model confidence scores. As Table~\ref{table:ece_grouped} displays, calibration had a high impact only on the multi-perspective models in which there is a big error difference (ECE) between the uncalibrated and calibrated counterparts. For instance, the multi-perspective BERT-large on HD has the model 
confidence scores of $0.29$ and $0.46$, while the LLMD version of the same model has $0.30$ and $0.46$ with and without calibration respectively. Similar results apply to the multi-perspective RoBERTa-large.

\newpage
\section{Conclusion \& Future Work}

In this work, we present a pipeline for integrating multi-perspective models into stance detection task on controversial topics, adaptable to various subjective applications. To promote responsibility in NLP systems, we advocate for the implementation of perspectivist models and sharing of disaggregated datasets. We believe these strategies are essential for more inclusive models and for the advancement of this emergent research area in NLP field.
We extended previous research by augmenting an existing dataset with summaries using state-of-the-art LLM and open-source LLM-based annotations to capture and preserve diverse viewpoints. We fine-tuned BERT-large and RoBERTa-large models using two methodologies: a baseline approach with hard aggregated labels and a novel approach with multi-perspective soft labels. 
Results show that multi-perspective models achieve better performances than baseline, with soft labels enhancing hard metrics (i.e. accuracy, precision, recall).
However, beyond improving model performance, we also applied calibration on model predictions to properly use them as model confidence scores, then employ these confidence values to compare the baseline and multi-
\vspace{0.5em}
perspective approaches. 

This work has potential limitations. Discarding instances without a majority label decreased the dataset size but made our experiments feasible for the comparative evaluation with the baseline model. 
Nonetheless, we believe that those instances are a valuable source for analyzing the multi-perspective approach which aims to learn from diverse perspectives instead of treating them as noise.  
One other possible solution is to increase the number of classes and annotators for each dataset instance, which we plan to pursue in future work. 
In this study, we used LLMs to gather annotations from different perspectives. However, previous research has shown that these models, when used as annotators, do not always align with human label distributions (opinions) accurately \cite{pavlovic2024effectiveness}. Specifically, state-of-the-art models like GPT-4 tend to be biased towards false positives, often labeling samples as offensive, abusive, or misogynistic. We recommend being mindful of such models' being inherently biased toward particular perspectives.
Apart from these, our analysis was constrained by computational resources, affecting batch size and model capacity.
In the future, we plan to expand our analysis by applying the current pipeline to a broader range of subjective tasks and datasets. We also aim to increase the number of baselines, following \cite{davani2022dealing}, with which we test the effectiveness of our method,in order to make the framework more generalizable. Moreover, we aim to explore deeper confidence-based results using soft labels, which can provide more nuanced insights into model uncertainty and prediction confidence.


\newpage

\appendix
\label{sec:applabels}

\begin{table}[h!]
\centering

\begin{subtable}[t]{0.48\textwidth}
\centering
\caption{Pegasus-CNN-DailyMail}
\scalebox{1.0}{
\begin{adjustbox}{width=\textwidth}
\begin{tabular}{lcc}
\toprule
\textbf{Metric} & \textbf{Chunk vs Summary} & \textbf{Doc vs Summary} \\
\midrule
\textbf{ROUGE-1} & & \\
Recall    & 0.15 & 0.15 \\
Precision & 0.92 & 0.92 \\
F1 Score  & 0.25 & 0.25 \\
\midrule
\textbf{ROUGE-2} & & \\
Recall    & 0.11 & 0.10 \\
Precision & 0.85 & 0.84 \\
F1 Score  & 0.18 & 0.17 \\
\midrule
\textbf{ROUGE-L} & & \\
Recall    & 0.15 & 0.15 \\
Precision & 0.92 & 0.91 \\
F1 Score  & 0.25 & 0.25 \\
\bottomrule
\end{tabular}
\end{adjustbox}
}
\label{tab:rouge_scores1}
\end{subtable}%

\begin{subtable}[t]{0.48\textwidth} 
\centering
\caption{BART-large-CNN}
\scalebox{1.0}{
\begin{adjustbox}{width=\textwidth} 
\begin{tabular}{lcc}
\toprule
\textbf{Metric} & \textbf{Chunk vs Summary} & \textbf{Doc vs Summary} \\
\midrule
\textbf{ROUGE-1} & & \\
Recall    & 0.23 & 0.25 \\
Precision & 0.95 & 0.93 \\
F1 Score  & 0.36 & 0.38 \\
\midrule
\textbf{ROUGE-2} & & \\
Recall    & 0.18 & 0.17 \\
Precision & 0.86 & 0.82 \\
F1 Score  & 0.28 & 0.28 \\
\midrule
\textbf{ROUGE-L} & & \\
Recall    & 0.23 & 0.25 \\
Precision & 0.95 & 0.92 \\
F1 Score  & 0.36 & 0.38 \\
\bottomrule
\end{tabular}
\end{adjustbox}%
}
\label{tab:rouge_scores2}
\end{subtable}%

\begin{subtable}[t]{0.48\textwidth} 
\centering
\caption{Falcon-7B-Instruct}
\scalebox{1.0}{
\begin{adjustbox}{width=\textwidth} 
\begin{tabular}{lcc}
\toprule
\textbf{Metric} & \textbf{Chunk vs Summary} & \textbf{Doc vs Summary} \\
\midrule
\textbf{ROUGE-1} & & \\
Recall    & 0.23 & 0.25 \\
Precision & 0.95 & 0.93 \\
F1 Score  & 0.36 & 0.38 \\
\midrule
\textbf{ROUGE-2} & & \\
Recall    & 0.18 & 0.17 \\
Precision & 0.86 & 0.82 \\
F1 Score  & 0.28 & 0.28 \\
\midrule
\textbf{ROUGE-L} & & \\
Recall    & 0.23 & 0.25 \\
Precision & 0.95 & 0.92 \\
F1 Score  & 0.36 & 0.38 \\
\bottomrule
\end{tabular}
\end{adjustbox}%
}
\label{tab:rouge_scores3}
\end{subtable}
\vspace{1.5em}
\caption{Comparison of ROUGE scores between different models.}
\end{table}

\bibliographystyle{vancouver}

\begin{thebibliography}{10}

\bibitem{sandri2023don}
Sandri M, Leonardelli E, Tonelli S, Je{\v{z}}ek E.
\newblock Why don’t you do it right? analysing annotators’ disagreement in subjective tasks.
\newblock In: Proceedings of the 17th Conference of the European Chapter of the Association for Computational Linguistics; 2023. p. 2428-41.

\bibitem{davani2024d3code}
Davani AM, D{\'\i}az M, Baker D, Prabhakaran V.
\newblock D3CODE: Disentangling Disagreements in Data across Cultures on Offensiveness Detection and Evaluation.
\newblock arXiv preprint arXiv:240410857. 2024.

\bibitem{santurkar2023whose}
Santurkar S, Durmus E, Ladhak F, Lee C, Liang P, Hashimoto T.
\newblock Whose opinions do language models reflect?
\newblock In: International Conference on Machine Learning. PMLR; 2023. p. 29971-30004.

\bibitem{zhi2024beyond}
Zhi-Xuan T, Carroll M, Franklin M, Ashton H.
\newblock Beyond preferences in ai alignment.
\newblock Philosophical Studies. 2024:1-51.

\bibitem{muscato2024overview}
Muscato B, Mala CS, Manerba MM, Gezici G, Giannotti F.
\newblock An Overview of Recent Approaches to Enable Diversity in Large Language Models through Aligning with Human Perspectives.
\newblock In: Proceedings of the 3rd Workshop on Perspectivist Approaches to NLP (NLPerspectives)@ LREC-COLING 2024; 2024. p. 49-55.

\bibitem{basile2021toward}
Basile V, Cabitza F, Campagner A, Fell M.
\newblock Toward a perspectivist turn in ground truthing for predictive computing.
\newblock arXiv preprint arXiv:210904270. 2021.

\bibitem{uma2021learning}
Uma AN, Fornaciari T, Hovy D, Paun S, Plank B, Poesio M.
\newblock Learning from disagreement: A survey.
\newblock Journal of Artificial Intelligence Research. 2021;72:1385-470.

\bibitem{akhtar2021whose}
Akhtar S, Basile V, Patti V.
\newblock Whose opinions matter? perspective-aware models to identify opinions of hate speech victims in abusive language detection.
\newblock arXiv preprint arXiv:210615896. 2021.

\bibitem{gezici2021evaluation}
Gezici G, Lipani A, Saygin Y, Yilmaz E.
\newblock Evaluation metrics for measuring bias in search engine results.
\newblock Information Retrieval Journal. 2021;24:85-113.

\bibitem{basile2020s}
Basile V, et~al.
\newblock It’s the end of the gold standard as we know it. on the impact of pre-aggregation on the evaluation of highly subjective tasks.
\newblock In: CEUR workshop proceedings. vol. 2776. CEUR-WS; 2020. p. 31-40.

\bibitem{leonardelli2023semeval}
Leonardelli E, Uma A, Abercrombie G, Almanea D, Basile V, Fornaciari T, et~al.
\newblock SemEval-2023 task 11: Learning with disagreements (LeWiDi).
\newblock arXiv preprint arXiv:230414803. 2023.

\bibitem{peterson2019human}
Peterson JC, Battleday RM, Griffiths TL, Russakovsky O.
\newblock Human uncertainty makes classification more robust.
\newblock In: Proceedings of the IEEE/CVF international conference on computer vision; 2019. p. 9617-26.

\bibitem{collins2022eliciting}
Collins KM, Bhatt U, Weller A.
\newblock Eliciting and learning with soft labels from every annotator.
\newblock In: Proceedings of the AAAI conference on human computation and crowdsourcing. vol.~10; 2022. p. 40-52.

\bibitem{pereyra2017regularizing}
Pereyra G, Tucker G, Chorowski J, Kaiser {\L}, Hinton G.
\newblock Regularizing neural networks by penalizing confident output distributions.
\newblock arXiv preprint arXiv:170106548. 2017.

\bibitem{baan2023uncertainty}
Baan J, Daheim N, Ilia E, Ulmer D, Li HS, Fern{\'a}ndez R, et~al.
\newblock Uncertainty in natural language generation: From theory to applications.
\newblock arXiv preprint arXiv:230715703. 2023.

\bibitem{hu2023uncertainty}
Hu M, Zhang Z, Zhao S, Huang M, Wu B.
\newblock Uncertainty in natural language processing: Sources, quantification, and applications.
\newblock arXiv preprint arXiv:230604459. 2023.

\bibitem{chang2024survey}
Chang Y, Wang X, Wang J, Wu Y, Yang L, Zhu K, et~al.
\newblock A survey on evaluation of large language models.
\newblock ACM Transactions on Intelligent Systems and Technology. 2024;15(3):1-45.

\bibitem{riccardi2023two}
Riccardi N, Desai RH.
\newblock The two word test: A semantic benchmark for large language models.
\newblock arXiv preprint arXiv:230604610. 2023.

\bibitem{gilardi2023chatgpt}
Gilardi F, Alizadeh M, Kubli M.
\newblock ChatGPT outperforms crowd workers for text-annotation tasks.
\newblock Proceedings of the National Academy of Sciences. 2023;120(30):e2305016120.

\bibitem{mohta2023large}
Mohta J, Ak K, Xu Y, Shen M.
\newblock Are large language models good annotators?
\newblock In: Proceedings on. PMLR; 2023. p. 38-48.

\bibitem{achiam2023gpt}
Achiam J, Adler S, Agarwal S, Ahmad L, Akkaya I, Aleman FL, et~al.
\newblock Gpt-4 technical report.
\newblock arXiv preprint arXiv:230308774. 2023.

\bibitem{dubey2024llama}
Dubey A, Jauhri A, Pandey A, Kadian A, Al-Dahle A, Letman A, et~al.
\newblock The llama 3 herd of models.
\newblock arXiv preprint arXiv:240721783. 2024.

\bibitem{pavlovic2024effectiveness}
Pavlovic M, Poesio M.
\newblock The Effectiveness of LLMs as Annotators: A Comparative Overview and Empirical Analysis of Direct Representation.
\newblock arXiv preprint arXiv:240501299. 2024.

\bibitem{zhu2023can}
Zhu Y, Zhang P, Haq EU, Hui P, Tyson G.
\newblock Can chatgpt reproduce human-generated labels? a study of social computing tasks.
\newblock arXiv preprint arXiv:230410145. 2023.

\bibitem{ziems2024can}
Ziems C, Held W, Shaikh O, Chen J, Zhang Z, Yang D.
\newblock Can large language models transform computational social science?
\newblock Computational Linguistics. 2024;50(1):237-91.

\bibitem{lee2023can}
Lee N, An NM, Thorne J.
\newblock Can Large Language Models Capture Dissenting Human Voices?
\newblock arXiv preprint arXiv:230513788. 2023.

\bibitem{wang2024human}
Wang X, Kim H, Rahman S, Mitra K, Miao Z.
\newblock Human-LLM collaborative annotation through effective verification of LLM labels.
\newblock In: Proceedings of the CHI Conference on Human Factors in Computing Systems; 2024. p. 1-21.

\bibitem{uma2020case}
Uma A, Fornaciari T, Hovy D, Paun S, Plank B, Poesio M.
\newblock A case for soft loss functions.
\newblock In: Proceedings of the AAAI Conference on Human Computation and Crowdsourcing. vol.~8; 2020. p. 173-7.

\bibitem{plank2022problem}
Plank B.
\newblock The'Problem'of Human Label Variation: On Ground Truth in Data, Modeling and Evaluation.
\newblock arXiv preprint arXiv:221102570. 2022.

\bibitem{devlin2018bert}
Devlin J.
\newblock Bert: Pre-training of deep bidirectional transformers for language understanding.
\newblock arXiv preprint arXiv:181004805. 2018.

\bibitem{liu2019roberta}
Liu Y.
\newblock Roberta: A robustly optimized bert pretraining approach.
\newblock arXiv preprint arXiv:190711692. 2019.

\bibitem{lin-2004-rouge}
Lin CY.
\newblock {ROUGE}: A Package for Automatic Evaluation of Summaries.
\newblock In: Text Summarization Branches Out. Barcelona, Spain: Association for Computational Linguistics; 2004. p. 74-81.
\newblock Available from: \url{https://aclanthology.org/W04-1013}.

\bibitem{zhang2019bertscore}
Zhang T, Kishore V, Wu F, Weinberger KQ, Artzi Y.
\newblock Bertscore: Evaluating text generation with bert.
\newblock arXiv preprint arXiv:190409675. 2019.

\bibitem{papineni2002bleu}
Papineni K, Roukos S, Ward T, Zhu WJ.
\newblock Bleu: a method for automatic evaluation of machine translation.
\newblock In: Proceedings of the 40th annual meeting of the Association for Computational Linguistics; 2002. p. 311-8.

\bibitem{jiang2023mistral}
Jiang AQ, Sablayrolles A, Mensch A, Bamford C, Chaplot DS, Casas Ddl, et~al.
\newblock Mistral 7B.
\newblock arXiv preprint arXiv:231006825. 2023.

\bibitem{groeneveld2024olmo}
Groeneveld D, Beltagy I, Walsh P, Bhagia A, Kinney R, Tafjord O, et~al.
\newblock Olmo: Accelerating the science of language models.
\newblock arXiv preprint arXiv:240200838. 2024.

\bibitem{guo2017calibration}
Guo C, Pleiss G, Sun Y, Weinberger KQ.
\newblock On calibration of modern neural networks.
\newblock In: International conference on machine learning. PMLR; 2017. p. 1321-30.

\bibitem{naeini2015obtaining}
Naeini MP, Cooper G, Hauskrecht M.
\newblock Obtaining well calibrated probabilities using bayesian binning.
\newblock In: Proceedings of the AAAI conference on artificial intelligence. vol.~29; 2015. .

\bibitem{davani2022dealing}
Davani AM, D{\'\i}az M, Prabhakaran V.
\newblock Dealing with disagreements: Looking beyond the majority vote in subjective annotations.
\newblock Transactions of the Association for Computational Linguistics. 2022;10:92-110.

\end{thebibliography}

\end{document}